\documentclass[runningheads]{llncs} 
\usepackage[T1]{fontenc}
\usepackage{graphicx}    
\usepackage{amsmath}
\usepackage{epsfig}
\usepackage{hyperref}
\begin{document}                            
\title{Evaluating and Benchmarking Foundation Models for Earth Observation and Geospatial AI}
\titlerunning{Evaluating and Benchmarking Foundation Models for Earth Observation}
%
%
\author{Nikolaos Dionelis$^1$, Casper Fibaek$^1$, Luke Camilleri$^{1,2}$, Andreas Luyts$^{1,3}$, Jente Bosmans$^1$, Bertrand Le Saux$^1$}
%
\authorrunning{N. Dionelis, C. Fibaek, et al., Submitted}    
%
%
\institute{$^1$ European Space Agency (ESA), ESRIN, $\Phi$-lab, Italy, $^2$ Trust Stamp, $^3$ VITO}
\maketitle                
%
%
\begin{abstract}                       
When we are primarily interested in solving several problems jointly with a given prescribed high performance accuracy for each target application, then Foundation Models should be used rather than problem-specific models. We focus on the specific vision application of Foundation Models for Earth Observation (EO) and geospatial AI. These models can solve important problems we are tackling, including for example land cover classification, crop type mapping, flood segmentation, building density estimation, and road regression segmentation. In this paper, we show that for a limited number of labelled data, Foundation Models achieve improved performance compared to problem-specific models. In this work, we also present our proposed evaluation benchmark for Foundation Models for EO. Benchmarking the generalization performance of Foundation Models is important as it has become difficult to standardize a fair comparison across the many different models. We present the results using our evaluation benchmark for EO Foundation Models and show that Foundation Models are label efficient in the downstream tasks and help us solve problems we are tackling in EO.
\keywords{Foundation Models for Earth monitoring \and Evaluation bench.}   
\end{abstract}
\section{Introduction}                       
An advantage of Foundation Models compared to problem-specific models is that for a limited number of labelled data, Foundation Models achieve improved performance.   
Label efficiency is important in real-world applications as for many use cases, both labelling and continuous re-labelling are needed.      
In the specific case of Earth Observation (EO) and remote sensing, labels \textit{change} over time.   
Also, data from satellites are \textit{unlabelled}.       
Annotating such data is difficult, requires expertise, and is costly in terms of time.       
An additional advantage of Foundation Models is that they perform \textit{sharing} across tasks and learn a common module, for example segmentation, needed for all the target applications we are trying to solve jointly with a given prescribed high performance accuracy for \textit{each} task.

The target applications of EO Foundation Models are important problems we are trying to \textit{solve}, such as land cover classification semantic segmentation, crop type mapping, and crop yield estimation.   
Additional target applications are flood segmentation, building density estimation, road \textit{regression} segmentation, estimation of the age of buildings, marine litter detection, methane plume segmentation, and change detection for wildfires, floods, and anomalies.        
Furthermore, there are also important EO problems that we would like to solve for which we have \textit{only} unlabelled data, i.e. no labels, for example iceberg detection.

\section{Solving $M$ tasks jointly with prescribed high accuracy}
Given a prescribed high performance for each task, e.g., accuracy $95\%$, we deal with $M$ problems \textit{jointly}.       
For EO Foundation Models, we address approximately $M=10$ target applications together.    
The prescribed high performance is crucial as we want the model to be useful; otherwise, people will \textit{not} use it.   
For Earth monitoring, we want generalization to a big geographical area/ large inference set.      
The performance stringent requirement drives everything.  
The \textit{two} alternatives are the following.        
For the use cases, for datasets D$1$, D$2$, ..., D$M$ that have labels, the alternative A is to perform supervised learning on the datasets. 
We name these tasks P$1$, P$2$, ..., P$M$.       
The alternative B is to perform \textit{self-supervised} learning on a common dataset $D'$. We name this task $L$. Then, we perform supervised learning for the target applications. We name these tasks Q$1$, Q$2$, ..., Q$M$. The dataset $D'$ contains relevant data, e.g., similar objects or data from the same satellite.   
The alternative A is using problem-specific models, solving \textit{each} problem on its own, and assuming the existence of a \textit{lot} of labels for each use case. The alternative B is using a common model and solving \textit{groups} of tasks that are of interest to us. Big common/ shared models are Foundation Models.      
For the alternative A, problem-specific models do \textit{not} have label efficiency: for limited labelled data, they yield \textit{low} performance accuracy (or F1-score or Intersection over Union (IoU)).  
There is no sample efficiency for these models and we have to pay too much and \textit{wait} too long for the labels. 
The performance requirement drives everything as the data size mainly depends on the prescribed high accuracy. 
The relationship between the size of the data and the accuracy is approximately linear. 
We cannot escape the \textit{large} size of the dataset because of the performance stringent requirement.    
In EO, the data size is: some TBs. 
Using common/ shared models is \textit{beneficial}: we learn the common representations. There is sharing across tasks: we learn the commonality, i.e. the common operations (segmentation) for \textit{re-usability} and efficiency.      
For the alternative B, i.e. for common models and Foundation Models, $N\%$ of the labels are needed that would otherwise be required.     
For EO Foundation Models, $N \approx 20$ and \textit{even} $10$.                 
For the alternative A (problem-specific models), \textit{all} the labels are needed. 

For the alternative A, the cost C$1$ (which is also directly related to the data \textit{size} and how large the architecture needs to be as these three are \textit{similar}) is:
\begin{equation}\label{eq:eqmain1}
\text{C}1 = \text{P}1 + \text{P}2 + ... + \text{P}M \approx My\text{,}    
\end{equation}
where typically $M=10$ tasks and $y$ is the cost or data for one task. Because of the \textit{high} accuracy requirement, $y$ is large, e.g., $100000$.   
This is why for problem-specific models, the cost, as well as the data size and how large is the architecture, is \textit{times} the number of tasks.     
For $M=10$ use cases, for the alternative A, we have times $10$, i.e. $\text{C}1=10y$ from \eqref{eq:eqmain1}. Next, for the alternative B, the cost is:
\begin{equation}
\text{C}2 = L + \text{Q}1 + \text{Q}2 + ... + \text{Q}M \approx y + N\% \,y \,M = y \, (1 + N\% \, M)\text{.}  
\end{equation}
This scales better than C$1$, i.e. $\text{C}2=3y$.        
Overall, $\text{C}2 \approx 300000$, $\text{C}1 \approx 1M$, and $\text{C}2 < \text{C}1$.   
Big common models achieve label efficiency for both segments and semantics.      
Segment label efficiency refers to the segments and their \textit{shape}.       
For both segment and semantic label efficiency, in remote sensing, continuous re-labelling is needed as we live in a \textit{dynamic} world: Earth is ever-changing.          
Human annotators are needed, as well as expert knowledge.                  
Also, imperfect labels exist in EO, i.e. \textit{noisy} labels.          
$\text{C}1$ grows linearly with $M$, i.e. O($M$), while $\text{C}2$ \textit{grows} linearly with $N\% \, M$, i.e. O($N\% \, M$).              
Because of the accuracy requirement and the \textit{linear} relationship between the data size and the accuracy, for problem-specific models, we train $10$ models that are \textit{approximately} as large as $10$ Foundation Models, i.e. it is like training $10$ Foundation Models.                      
Also, for problem-specific models, a lot of labels are needed which are expensive in terms of \textit{both} cost and time.

\section{Our Proposed Evaluation Benchmark for FMs for EO}
\begin{figure} 
  \centering          
  \centerline{\epsfig{figure=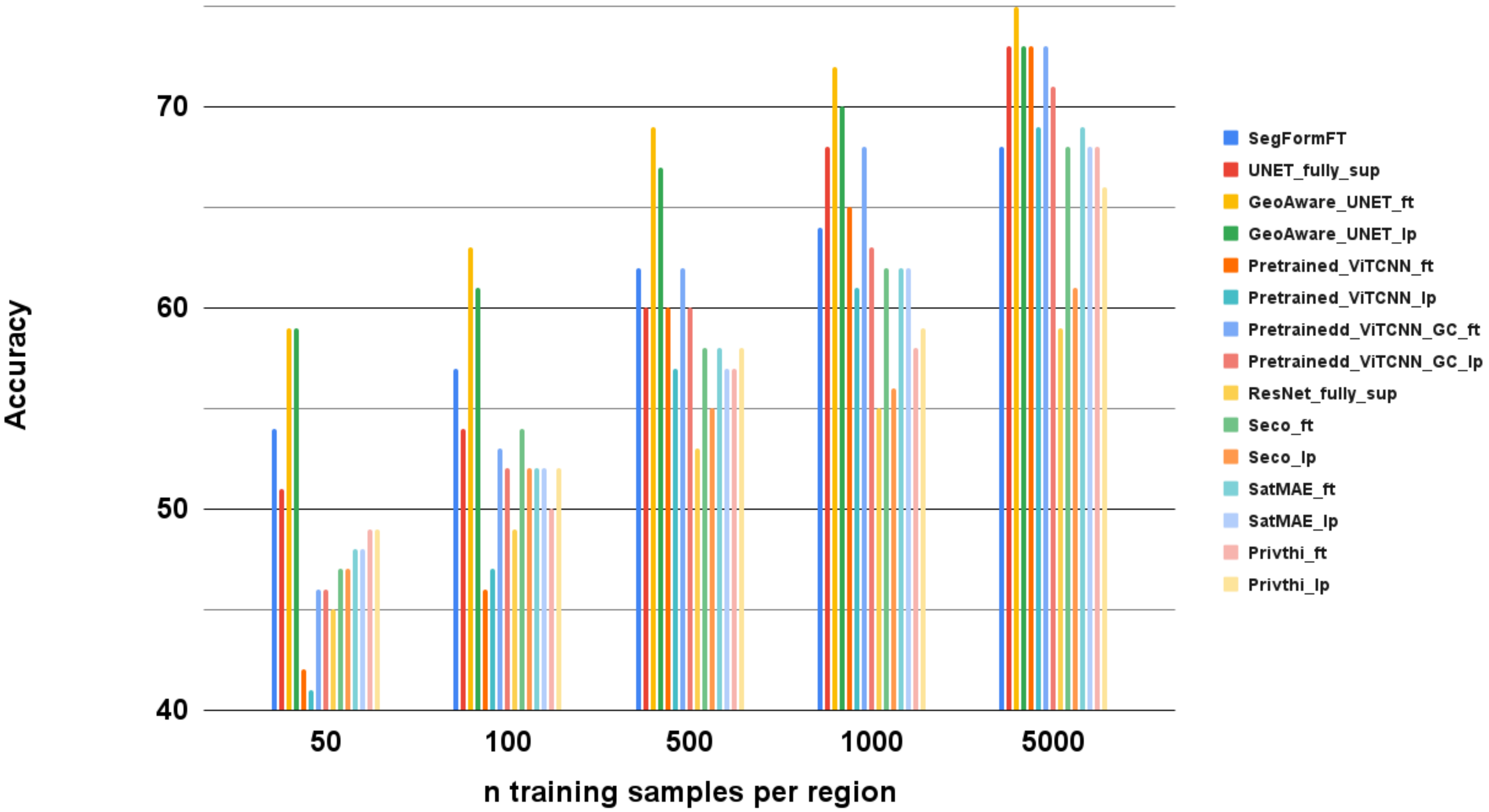,width=0.83\textwidth}}             
\vspace{-8pt}
\caption{Evaluating, \textit{benchmarking}, and ranking Foundation Models for EO and geospatial AI on the downstream task of semantic segmentation land cover classification.} \label{fig:figure1}                
\end{figure}

Evaluating and benchmarking Foundation Models in terms of their generalization performance is important as it has become increasingly difficult to standardize a \textit{fair} comparison across the many different models.               
For the specific vision application of Foundation Models for EO and geospatial AI \cite{ref_proc1,ref_proc2,ref_proc3}, we present our proposed evaluation benchmark and show that for a \textit{limited} number of labelled data, Foundation Models achieve improved results compared to problem-specific models.      
Foundation Models are label efficient in the downstream tasks \cite{PhilEO2023,PhilEOEGU}.            
For semantic segmentation land cover classification (lc), the evaluation results are presented in Fig.~\ref{fig:figure1}.      
We examine both \textit{fine-tuning} (ft) and linear probing (lp).%
%
%

\noindent Geo-location classification pre-training is used for the models that we have developed in-house.                        
These are the \textit{geo-aware} models in Fig.~\ref{fig:figure1}.      
As a pre-text task, our Foundation Model Version 1.0 performs longitude and latitude satellite metadata information learning.     
For this, we have used a \textit{global} unlabelled dataset of satellite Sentinel-2 L2A data and $10$ spectral bands.   
As a downstream task, we perform fine-tuning (or linear probing) on the labelled dataset WorldCover\footnote{\href{http://worldcover2020.esa.int/data/docs/WorldCover_PUM_V1.1.pdf}{http://worldcover2020.esa.int/data/docs/WorldCover$\_$PUM$\_$V1.1.pdf}}.           
According to the results in Fig.~\ref{fig:figure1}, the percentage \textit{improvement} of Foundation Models compared to problem-specific models is approximately $18.52\%$ when there are limited samples of labelled data, e.g., $100$ images per region (geo-aware U-Net ft and U-Net fully-supervised).        
We have examined both a \textit{Transformer}-based architecture, i.e. Vision Transformer (ViT), and a U-Net-based architecture.

For the task of estimating the label at the image level (rather than at the \textit{pixel} level) for land cover classification, according to our results, the percentage improvement of Foundation Models compared to problem-specific models is approximately $16.36\%$ when limited labels are used, e.g., $100$ samples per region (geo-aware U-Net ft vs. U-Net fully-supervised, $0.64$ and $0.55$ respectively).

\begin{figure}
  \centering      
  \centerline{\epsfig{figure=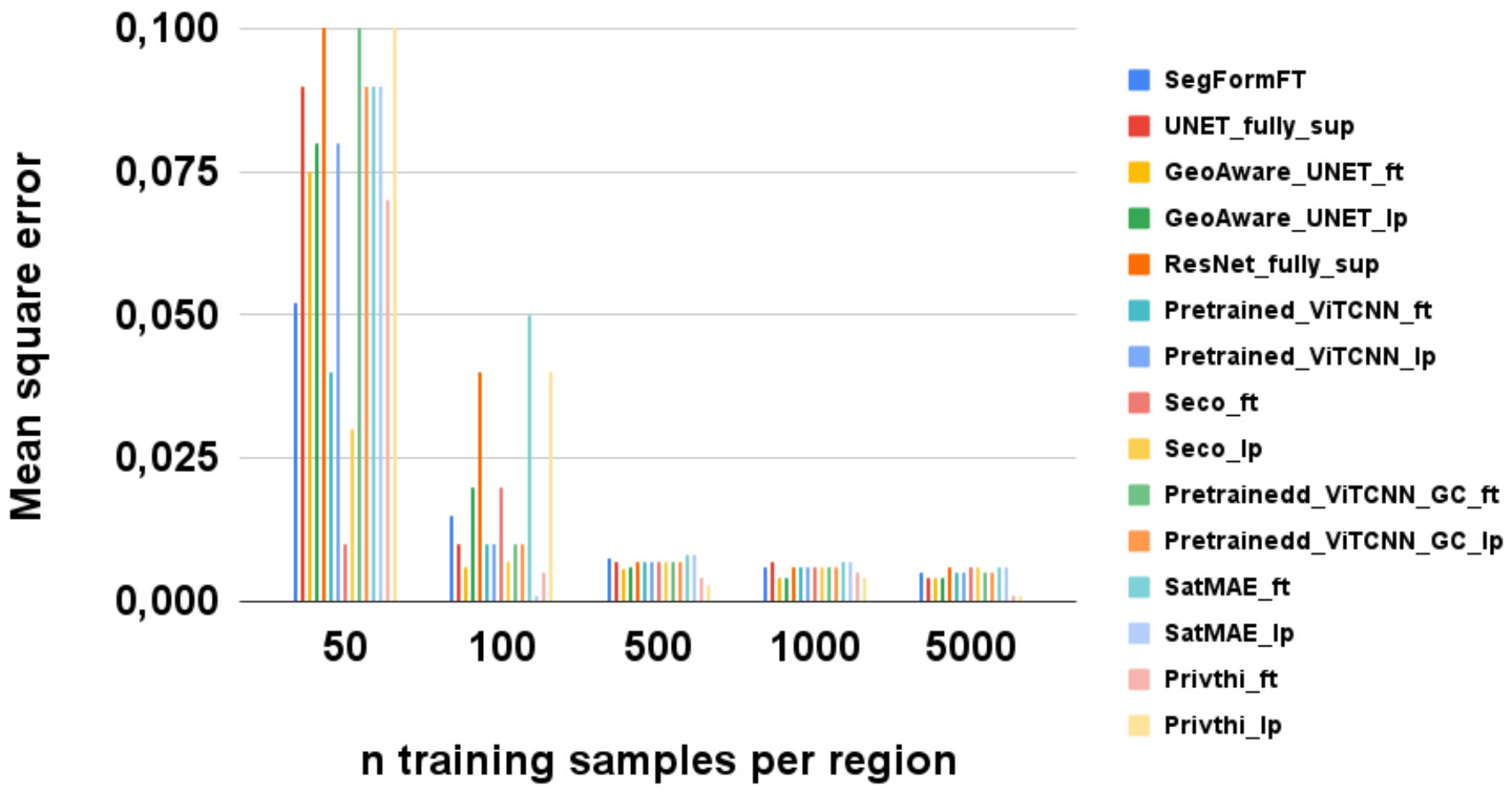,width=0.6465\textwidth}}         
\vspace{-8pt}\caption{Evaluation of Foundation Models for EO on the target application of estimating how dense and \textit{close} to each other buildings are, in the MSE metric (regression task).} \label{fig:figure2}    
\end{figure}

Next, for the task of estimating how dense and close to each other buildings are, the results are presented in Fig.~\ref{fig:figure2}.            
For this \textit{regression} downstream task, the evaluation metric is the Mean Squared Error (MSE).     
We compare $15$ models in total.  
For this specific use case, the percentage improvement of Foundation Models \textit{compared} to problem-specific models is $86\%$ when there are limited labelled data: $100$ samples per region (geo-aware U-Net and U-Net fully-supervised).%
%
%
\section{Conclusion}                              
To solve several problems jointly with a prescribed high accuracy for each task, we use Foundation Models. For the vision application of Foundation Models for EO, for limited labelled data, Foundation Models outperform problem-specific models in our proposed evaluation benchmark for Foundation Models for EO.










%
%
%
%

\end{document}